\documentclass[sigconf]{acmart}

\usepackage{multirow,array, balance}
\newcommand{\PreserveBackslash}[1]{\let\temp=\\#1\let\\=\temp}
\newcolumntype{C}[1]{>{\PreserveBackslash\centering}p{#1}}
\newcolumntype{R}[1]{>{\PreserveBackslash\raggedleft}p{#1}}
\newcolumntype{L}[1]{>{\raggedright\let\newline\\\arraybackslash\hspace{0pt}}m{#1}}

\AtBeginDocument{%
  }

\copyrightyear{2025}
\acmYear{2025}
\setcopyright{acmlicensed}
\acmConference[SIGIR '25] {Proceedings of the 48th International ACM SIGIR Conference on Research and Development in Information Retrieval}{ July 13--18, 2025}{Padua, Italy.}
\acmBooktitle{Proceedings of the 48th International ACM SIGIR Conference on Research and Development in Information Retrieval (SIGIR '25), July 13--18, 2025, Padua, Italy}
\acmISBN{979-8-4007-1592-1/25/07}
\acmDOI{10.1145/3726302.3730164}
\begin{document}

\title{Are Information Retrieval Approaches Good at Harmonising Longitudinal Survey Questions in Social Science?}

\author{Wing Yan Li}
\authornote{Corresponding author}
\affiliation{%
  \institution{University of Surrey}
  \country{United Kingdom}
  }
  
\email{wingyan.li@surrey.ac.uk}

\author{Zeqiang Wang}
\affiliation{%
  \institution{University of Surrey}
  \country{United Kingdom}
  }
\email{zeqiang.wang@surrey.ac.uk}

\author{Jon Johnson}
\affiliation{%
  \institution{University College London}
  \country{United Kingdom}
  }
\email{jon.johnson@ucl.ac.uk}

\author{Suparna De}
\affiliation{%
  \institution{University of Surrey}
  \country{United Kingdom}
  }
\email{s.de@surrey.ac.uk}

\renewcommand{\shortauthors}{Wing Yan Li, Zeqiang Wang, Jon Johnson, and Suparna De}

\begin{abstract}
Automated detection of semantically equivalent questions in longitudinal social science surveys is crucial for long-term studies informing empirical research in the social, economic, and health sciences. Retrieving equivalent questions faces dual challenges: inconsistent representation of theoretical constructs (i.e.~concept/sub-concept) across studies as well as between question and response options, and the evolution of vocabulary and structure in longitudinal text. To address these challenges, our multi-disciplinary collaboration of computer scientists and survey specialists presents a new information retrieval (IR) task of identifying concept (e.g.~Housing, Job, etc.) equivalence across question and response options to harmonise longitudinal population studies. This paper investigates multiple \textbf{unsupervised} approaches on a survey dataset spanning 1946-2020, including probabilistic models, linear probing of language models, and pre-trained neural networks specialised for IR. We show that IR-specialised neural models achieve the highest overall performance with other approaches performing comparably. Additionally, the re-ranking of the probabilistic model's results with neural models only introduces modest improvements of 0.07 at most in F1-score. Qualitative post-hoc evaluation by survey specialists shows that models generally have a low sensitivity to questions with high lexical overlap, particularly in cases where sub-concepts are mismatched. Altogether, our analysis serves to further research on harmonising longitudinal studies in social science.

\end{abstract}




\maketitle

\section{Introduction}

Large-scale longitudinal social surveys are fundamental to understanding societal changes across generations. 
Through questions tailored to specific populations and temporal contexts, surveys capture theoretical constructs, also referred to as concept or sub-concept, (e.g. housing conditions).
Shifts in questionnaire structure, semantic frameworks, and concept usage due to shifting social contexts and target respondent characteristics (e.g. education level) mean that surveys vary in language, structure and response options.
This poses a significant challenge when harmonising studies that span different time periods and populations.

\begin{table}[t]
\small
\centering
\caption{Typology of survey questions with the corresponding \textcolor{blue}{concepts} and/or \textcolor{orange}{sub-concepts} highlighted. \textbf{Bold} texts are questions and \textit{italic} texts are response options.}
\label{tab:q_typo}
    \begin{tabular}{C{0.1\textwidth}|L{0.33\textwidth}}
    \toprule
        Question Type & \multicolumn{1}{c}{Example} \\ \midrule
        Standard & \textbf{What is his \textcolor{blue}{current job}?} \newline \textcolor{orange}{\textit{1. self-employed | 2. manager | 3. foreman} \newline \textit{4. employee}}\\ \midrule
        Qualified & \textbf{The thought of \textcolor{blue}{harming myself} has occurred to me:} \newline \textit{1. yes, quite often | 2. sometimes | 3. hardly ever | 4. never} \\ \midrule
        Compound & \textbf{What is she \textcolor{blue}{allergic} to? (tick all that apply) \textcolor{orange}{dog}} \newline \textit{1. yes} \\
    \bottomrule
    \end{tabular}
\end{table}

In this work, we address and investigate two main challenges when detecting conceptually equivalent questions in longitudinal surveys:
(1) The linguistic challenge of capturing semantic equivalence despite surface-level variations in vocabulary and syntax. For instance, ``dwelling'' may evolve to ``household'' over time while maintaining the same underlying concept. 
(2) The structural challenge in which equivalent concepts may be distributed differently between question text and response options.
In Table~\ref{tab:q_typo}, the sub-concept in the response categories of the \textit{standard} question could fundamentally alter the concept being measured.
Conversely, \textit{qualified} questions do not capture any sub-concepts.

While information retrieval (IR) techniques have advanced significantly in recent years, their applicability to survey harmonisation remains under explored. Traditional approaches like BM25~\cite{robertson2009probabilistic} excel at lexical matching but may miss semantic equivalences obscured by terminology shifts. Contemporary neural models~\cite{karpukhin2020dense,cedr,colbert,colbert-x} while powerful at semantic modelling risk overfitting to superficial patterns without capturing the domain-specific nature of survey concepts. Moreover, existing work rarely addresses the unique requirements of structured survey questions where both question text and coded response options jointly define the measured construct. In this collaboration between computer scientists and survey specialists, 
we bridge this gap through three main contributions:
\begin{enumerate}
    \item We formulate the task of detecting conceptual equivalence in longitudinal surveys as an IR problem, specifically addressing the challenge of integrating question and response text as coherent input sequences.
    \item We present a systematic evaluation of unsupervised IR approaches on a dataset spanning 75 years (1946-2020) by comparing probabilistic models (BM25), linearly probed language models (DeBERTa, Qwen), and IR-specialised neural networks (M3-Embedding).
    \item We combine quantitative metrics based on topic code alignment with qualitative analysis by domain experts to reveal both the promise and limitations of current IR methods for survey harmonisation.
\end{enumerate}

Our results reveal that while IR-optimised neural models achieve the highest overall performance ($F1=0.79$), traditional approaches like BM25 remain competitive ($F1=0.75$). Notably, qualitative analysis reveals that all models struggle with questions showing high lexical overlap but divergent sub-concepts, suggesting the need for more sophisticated approaches to concept granularity.


\section{Related Work}

\paragraph{\textbf{Computational Social Science}}
Survey data harmonisation is the process of making different surveys more comparable and consistent across various variables such as studies, concepts, and time.
Existing approaches rely heavily on classification schemas that are manually annotated by experts. These methods are typically statistical or procedural based~\cite{dubrow2016rise}, thus leading to challenges (e.g.~language barriers) when conducting cross-cultural research~\cite{tsai2024challenges}.
On the other hand,~\cite{stohr2024advancing} proposed to train domain-specific Word2Vec embeddings using social science literature. Their evaluation of various tasks such as text classification with data collected from a sociology dictionary showed promising results.

\paragraph{\textbf{Information Retrieval Models}}
Traditional IR continuous to adopt probabilistic-based methods that are based on statistical evidence (e.g.~term-frequencies~\cite{robertson2009probabilistic,salton1988term}).
Recently, neural IR models have garnered increasing attention by performing searches based on the underlying semantic information. Pre-trained language models like DeBERTa~\cite{he2021deberta} are commonly adopted as the reader model to embed sequences for a downstream retrieval model~\cite{cedr,colbert,colbert-x}.
Some works also proposed IR-specialised models with various training strategies \cite{guo2025came,li2023making} and architectural designs \cite{xiong2020approximate,chen-etal-2024-m3} that aim at getting the best of both worlds (i.e.~syntactic and semantic features).

\section{Experiments}

Suppose there are $n$ sets of questionnaires $S={\{S_1,S_2,...,S_n\}}$.
Let $U=\bigcup_{i=1}^{n}{S_i}$ be the union of all questionnaires.
For each query question and the corresponding response options $(q,r)\in{U}$, the objective of the experiment is to identify all potential \textbf{\textit{conceptually equivalent}} in $U-\{(q,r)\}$.

\paragraph{\textbf{Dataset}}
We use questions from a large-scale longitudinal social science survey project consisting of 318 questionnaires. The questions possess diverse longitudinal spans ranging from 1946 to 2020. In total, there are 42,161 questions with three different typologies: \textit{standard, qualified}, and \textit{compound} questions, as illustrated in Table~\ref{tab:q_typo}.
The \textit{standard} questions have the most diverse distribution of concepts by presenting concepts and sub-concepts in the questions and response options, respectively.
This introduces an inherent complexity and additional challenge since \textit{standard} questions only appear as code list questions. 
As such, we consider code list questions\footnote{30,863 code list questions in total.} only and construct our input sequence as the concatenation of the question and response options.

\paragraph{\textbf{Implementation Details}}
In the following experiments, all models are \textit{\textbf{training-free}}.
Our baseline model is BM25\footnote{We use the bm25s library~\cite{bm25s} for efficient implementations}, a probabilistic model which retrieves equivalent items based on term frequencies. We employ linear probing on pre-trained LLMs as reader models to embed input sequences. Following the bi-encoder paradigm, inputs (i.e.~questions) are encoded independently.
Various encoder models including the SBERT\footnote{The chosen base model is \texttt{multi-qa-mpnet-base-dot-v1}.}~\cite{reimers-gurevych-2019-sentence} sentence transformer, DeBERTa-v3~\cite{he2021deberta}, and pre-trained decoder model like Qwen\footnote{We use Qwen-2.5 with 3B parameters and 4-bit quantisation.}~\cite{yang2024qwen2} are considered.
For efficiency, we follow the late interaction framework~\cite{colbert,colbert-x} where representations are pre-computed offline with cosine distance ($1-cosine\_similarity$) as the similarity measure.

In addition, we investigate different ways of generating sequence representations in the DeBERTa and Qwen models.
\textbf{(1) Mean-pooled sequence representation (mean-rep).} The mean of all the token representations from the final encoder or decoder layer is taken.
\textbf{(2) Sequence summary token (SST-rep).} For encoder models, we use the [CLS] representation. For decoder models, we use the last token representation due to the autoregressive nature of decoder models.
The chosen SBERT model uses [CLS] pooling for sequence representation.
Since the model is trained end-to-end using [CLS] pooling, we do not investigate the impact of using mean-rep as this will result in sub-optimal performance.

For the IR-specialised model, we utilise M3-Embedding~\cite{chen-etal-2024-m3}, a multi-lingual, multi-granularity, and multi-functionality embedding model designed for IR tasks. It is trained with self‑knowledge distillation to harmonise dense, sparse (lexical), and multi-vector retrieval signals within a single architecture. The backbone model is a pre-trained XLM-RoBERTa~\cite{conneau-etal-2020-unsupervised}. During retrieval, it uses mean-rep and SST-rep to capture both word ordering and context nuances.

Specifically, we compare two variants of M3-Embedding, \texttt{BGE-m3} and \texttt{BGE-reranker-m3}. The former follows the bi-encoder paradigm while the latter follows the cross-encoder paradigm where query questions and the corresponding candidate question are encoded in pairs. \texttt{BGE-m3} is leveraged for both end-to-end ranking and re-ranking task while \texttt{BGE-reranker-m3} is used for the re-ranking task only.
In both models, the score function is the weighted\footnote{We follow the hyper-parameter settings as suggested in~\cite{chen-etal-2024-m3}.} sum of the three retrieval scores mentioned above.

Lastly, for the re-ranking task, neural models are asked to refine the top-50 questions per query from BM25. Other than the \texttt{BGE-reranker-m3}, all other models follows the LLM linear probing setup as described above.

\paragraph{\textbf{Evaluation}}
Survey questions are tagged using a hierarchical label ontology with 16 top-level topics (e.g.~Education, Health) and 120 fine-grained sub-topics (e.g.~Education-Primary Schooling, Education-Higher Education). 
We evaluate model performance by comparing the top-level topic codes between the query question and its most equivalent match as identified by the model. This is based on the assumption that equivalent questions address the same general topic.
Note that there is no randomness in the experiments as our models are \textbf{training-free} (i.e.~possess fixed weights).


\section{Results}

\begin{table}[t]
\caption{Model performances evaluated by matching the top-level topic code of the query question and \textit{the most similar question} (i.e.~top-1) as returned per model. BM25 is the baseline. For DeBERTa and Qwen, \texttt{mean}, \texttt{[CLS]} and \texttt{last} indicates the type of sequence representation used.}
\label{tab:model_perf}
\centering
    \begin{tabular}{c|c|cccc}
    \toprule
        Model & Setup & Precision & Recall & F1 & Accuracy \\ \midrule
        \multicolumn{5}{l}{End-to-End Ranking} \\ \midrule \midrule
        \multicolumn{2}{c|}{BM25} & 0.75 & 0.75	& 0.75 & 0.80 \\\midrule
        \multicolumn{2}{c|}{SBERT} & 0.73 & 0.55 & 0.62 & 0.66 \\ 
        \multicolumn{2}{c|}{BGE-m3} & \textbf{0.80} & \textbf{0.79} & \textbf{0.79} & \textbf{0.84}\\\midrule
        \multirow{2}{*}{DeBERTa} & mean & 0.69 & 0.69 & 0.69 & 0.78 \\
        & \texttt{[CLS]} & 0.68 & 0.68 & 0.68 & 0.77 \\ \midrule
        \multirow{2}{*}{Qwen} & mean & 0.75 & 0.72 & 0.73 & 0.83 \\
        & last & 0.68 & 0.67 & 0.67 & 0.80 \\ \midrule
        
        \multicolumn{5}{l}{Re-Ranking} \\ \midrule \midrule
        \multicolumn{2}{c|}{SBERT} & 0.74 & 0.55 & 0.62 & 0.66 \\ 
        \multicolumn{2}{c|}{BGE-m3} & 0.77 & 0.78 & 0.77 & 0.83 \\ 
        \multicolumn{2}{c|}{BGE-reranker-m3} & 0.74 & 0.73 & 0.74 & 0.81 \\ \midrule
        \multirow{2}{*}{DeBERTa} & mean & 0.70 & 0.71 & 0.70 & 0.81 \\
        & \texttt{[CLS]} & 0.76 & 0.74 & 0.75 & 0.80 \\ \midrule
        \multirow{2}{*}{Qwen} & mean & 0.74 & 0.71 & 0.72 & 0.83 \\
        & last & 0.70 & 0.68 & 0.69 & 0.81 \\
    \bottomrule
    \end{tabular}
\end{table}

End-to-end ranking is more challenging than re-ranking as there are more questions to search for. Table~\ref{tab:model_perf} shows model performances evaluated by comparing the top-level topic code of the query question and the corresponding most similar question. Interestingly, \texttt{BGE-m3} achieved the highest performance metric across the board, outperforming \texttt{BM25} by 0.04 on average across all metrics. Potentially, this contributes to the integrated score function of \texttt{BGE-m3} that scores questions from multiple aspects. Generally, there is an average gap of 0.08 between F1-scores and accuracies, particularly pronounced in \texttt{Qwen-last}.
This consistent gap suggests that models are relatively weak at identifying positive cases (i.e.~relevant items) but good at identifying non-relevant items.

\paragraph{\textbf{Types of Representations}}
Models with mean-rep consistently outperform models with SST-rep across the board. Intuitively, averaging representations will lead to information loss since the operation smooths out the distributional information embedded in the representations. This disregards information such as word ordering and interaction between words~\cite{gupta-jaggi-2021-obtaining}. Therefore, mean-rep contain shallower semantic information than SST-rep. Potentially, this implies that embedding input sequences with relatively shallower semantic information might be more favourable for this task.

This is also found in the re-ranking results. Re-ranking \texttt{BM25} outputs with neural models is expected to enhance performance by compensating the lack of semantic information in \texttt{BM25}. However, results show little to no improvements compared to the baseline and end-to-end approaches suggesting that a balance between syntactic and semantic info seems to be critical for this task. 

\begin{table}[t]
\caption{Distribution of labels presented in percentages (\%). There are 203 randomly selected questions in total. The labels are:  1 --- exact match; 1a --- equivalent; 2 --- sub-concept mismatch, and 3 --- total mismatch.}
\label{tab:class_dist}
\centering
    \begin{tabular}{c|cccc}
    \toprule
        Model & 1 & 1a & 2 & 3 \\ \midrule
        BM25 & 65.02\%& 9.85\% & 13.30\% & 11.82\% \\ \midrule
        BGE-m3 & 67.98\% & 7.88\% & 13.79\% & 10.34\% \\
        DeBERTa-mean & 57.64\% & 7.39\% & 15.27\% & 19.70\% \\
        Qwen-mean & 66.50\% & 4.93\% & 14.29\% & 14.29\% \\
    \bottomrule
    \end{tabular}
\end{table}

\begin{table*}
\small
\caption{Selected qualitative examples from 203 random samples. \textbf{Bold texts} are questions and \textit{italic texts} are response options. ``Label'' indicates the type of relationship between the most similar question retrieved and query question, respectively.}
\label{tab:example}
    \begin{tabular}{L{0.04\textwidth}|L{0.21\textwidth}|L{0.22\textwidth}L{0.22\textwidth}L{0.21\textwidth}}
    \toprule
        Index & \multicolumn{1}{c|}{Query Question} & \multicolumn{1}{c}{BM25} & \multicolumn{1}{c}{BGE-m3} & \multicolumn{1}{c}{Qwen-mean} \\ \midrule
        
        \multirow{2}{*}{(a)} & (NCDS Age 33 Cohort Member Interview) \newline \textbf{And how satisfied or dissatisfied are you with the area you live in?} \newline \textit{1, very satisfied | 2, fairly satisfied | ... | 8, don't know}
        & (Wave 3 Questionnaire) \newline \textbf{On the whole, are you very satisfied, fairly satisfied, a little dissatisfied or very dissatisfied with the way democracy works in this country?} \newline \textit{1, very satisfied | ... | 4, very dissatisfied}
        & (1989 Main Questionnaire) \newline \textbf{How do you feel about living in this district? Would you say that you are} \newline \textit{7, very satisfied | ... | 1, very dissatisfied}
        & (1989 Main Questionnaire) \newline \textbf{And how do you feel about your present accommodation?} \newline \textit{7, very satisfied | ... | 1, very dissatisfied} \\ \cmidrule{2-5}

        & \multicolumn{1}{c|}{Label} & \multicolumn{1}{c}{3} & \multicolumn{1}{c}{1a} & \multicolumn{1}{c}{1a} \\ \midrule

        \multirow{2}{*}{(b)} & (Me and My Baby) \newline \textbf{What arrangements have you made about looking after your baby when you begin work? day nursery} \newline \textit{1, yes | 2, no | 9, don't know}
        & (Me and My Baby) \newline \textbf{What arrangements have you made about looking after your baby when you begin work? baby's grandparent} \newline \textit{1, yes | 2, no | 9, don't know}
        & (Me and My Baby) \newline \textbf{What arrangements have you made about looking after your baby when you begin work? other (please describe)} \newline \textit{1, yes | 2, no | 9, don't know} 
        & (Me and My Baby) \newline \textbf{What arrangements have you made about looking after your baby when you begin work? partner} \newline \textit{1, yes | 2, no | 9, don't know} \\ \cmidrule{2-5}

        & \multicolumn{1}{c|}{Label} & \multicolumn{1}{c}{2} & \multicolumn{1}{c}{2} & \multicolumn{1}{c}{2} \\ \midrule

        \multirow{2}{*}{(c)} & (Me and My School) \newline \textbf{Do you like answering questions in class?} \newline \textit{1, never | 2, sometimes | 3, often (most days) | 4, always}
        & (MCS Age 7 Cohort Member Paper Self-Completion) \newline \textbf{How much do you like answering questions in class?} \newline \textit{1, i like it a lot | 2, i like it a bit | 3, i don't like it}
        & (Me and My School) \newline \textbf{Do you feel happy at school?} \newline \textit{1, never | 2, sometimes | 3, often | 4, always}
        & (Me and My School) \newline \textbf{Do you get homework? }\newline \textit{1, never | 2, sometimes | 3, often | 4, always} \\ \cmidrule{2-5}
        
        & \multicolumn{1}{c|}{Label} & \multicolumn{1}{c}{1a} & \multicolumn{1}{c}{3} & \multicolumn{1}{c}{3} \\ \midrule

        \multirow{2}{*}{(d)} & (Your Health Events and Feelings) \newline \textbf{Do you use gas for cooking?} \newline \textit{1, yes, ring(s) only | ... | 4, no, not at all}
        & (Partner and Home) \newline \textbf{Do you use gas for cooking?}\newline \textit{1, yes, ring(s) only | ... | 4, no, not at all}
        & (Looking After the Baby) \newline \textbf{Do you use gas for cooking?} \newline \textit{1, yes, ring(s) only | ... | 4, no, not at all}
        & (Partner and Home) \newline \textbf{Do you use gas for cooking?} \newline \textit{1, yes, ring(s) only | ... | 4, no, not at all} \\ \cmidrule{2-5}

        & \multicolumn{1}{c|}{Label} & \multicolumn{1}{c}{1} & \multicolumn{1}{c}{1} & \multicolumn{1}{c}{1} \\
    \bottomrule
    \end{tabular}
\end{table*}

\paragraph{\textbf{Qualitative Analysis}}
Questions with the same topic code are assumed equivalent, but this may not always be accurate since concepts or sub-concepts are more granular and question-specific.
As such, 203 question pairs are randomly selected and then manually inspected by survey specialists.
After examining the query question and its corresponding most similar question, four labels are defined to indicate whether the pair is: 
1 --- exact match; 1a --- equivalent; 2 --- sub-concept mismatch, or 3 --- total mismatch. 

Here, we focus on the baseline and end-to-end ranking models, specifically, model variants with better performance on F1-score and accuracy, such as \texttt{BGE-m3} and \texttt{DeBERTa-mean}.
Table~\ref{tab:class_dist} shows the label distribution of the considered models. 
Whilst models perform well in identifying predominantly exact matching cases, class 2 and class 3 instances represent the next highest frequency in the distribution, with class 1a consistently being the smallest group.

The high proportion of exact matching cases in all models is likely due to the relatively low complexity of identifying such cases.
Moreover, the repetition of identical questions across different surveys, such as Table~\ref{tab:example}~(d), further contributes to the prevalence of class 1 instances.
Models generally identify more class 2 instances than class 3 instances.
This indicates that models are insensitive to subtle lexical variations that alter the underlying meaning, hence the theoretical constructs.
As demonstrated in Table~\ref{tab:example}~(b), question pairs labelled as sub-concept mismatches typically show high lexical overlap with minor differences in the syntax.

Class 1a instances are challenging to identify as they often involve questions that appear as a rephrasing of the query (Table~\ref{tab:example}~(a) and (c)). While one might expect that semantic understanding is required to detect such cases, our concatenation of the question and response options introduces a high degree of lexical overlap from the response list. This characteristic likely favours \texttt{BM25} that identifies the most class 1a instances compared to other models.

All in all, aligning with the previous claim, a balance between syntactic and semantic information is crucial for this task. Identifying conceptually equivalent questions extends beyond finding similar texts. Not only do models have to ``understand'' the underlying meaning but also be sensitive to subtle changes that could alter the concept or sub-concept of the question. 
This nuance underscores the importance of calibrating the emphasis on both surface-level patterns and deeper semantic aspects, respectively.

\section{Conclusion}

This paper presents a systematic evaluation of \textbf{unsupervised} information retrieval approaches for harmonising longitudinal survey questions in social science research.
We find that while IR-specialised models like \texttt{BGE-m3} achieve the highest overall performance ($F1=0.79$), traditional approaches such as \texttt{BM25} remain surprisingly competitive ($F1=0.75$). This suggests that lexical matching remains crucial for survey harmonisation even as semantic understanding improves. 
Our qualitative analysis further reveals that models generally excel at identifying exact (or equivalent) matches but struggle with subtle conceptual variations, particularly when sub-concepts differ despite high lexical overlap.
Secondly, our experiments with different representations indicate that balancing syntactic and semantic information is critical. The superior performance of mean-rep over SST-rep suggests that shallower semantic modelling might be more appropriate for survey question matching. This finding challenges the common assumption that deeper semantic understanding always leads to better performance in text matching tasks. 
Finally, our future work includes: (1) Hyper-parameter optimisation for domain adaptation. For \texttt{BGE-m3}, systematic hyper-parameter search may re-calibrate the balance between lexical and semantic matching signals for survey harmonisation.
(2) Fine-grained concept modelling. Hierarchical modelling approaches may better capture the granular relationships between survey concepts and their components. 
(3) Domain-specific language models. Developing language models specifically pre-trained on social science survey corpora may better capture domain-specific terminology and structure of the survey questions.




\bibliographystyle{ACM-Reference-Format}
\balance
\bibliography{custom}


\end{document}